\documentclass[conference]{IEEEtran}
\IEEEoverridecommandlockouts
\usepackage{cite}
\usepackage{amsmath,amssymb,amsfonts}
\usepackage{algorithmic}
\usepackage{graphicx}
\usepackage{textcomp}
\usepackage{xcolor}
\usepackage{listings}
\usepackage{tcolorbox}
\usepackage{booktabs}
\usepackage[table]{xcolor}
\usepackage{amssymb}
\usepackage{bbding}
\usepackage{graphicx}
\usepackage{subcaption}
\usepackage{diagbox}
\usepackage{comment}
\usepackage{hyperref}
\usepackage{fancyhdr}
\usepackage{eso-pic}

\def\BibTeX{{\rm B\kern-.05em{\sc i\kern-.025em b}\kern-.08em
    T\kern-.1667em\lower.7ex\hbox{E}\kern-.125emX}}
\begin{document}

\newcommand{\IEEEcopyrightbox}{
\AddToShipoutPictureFG*{
\put(57,33){
\begin{minipage}{500pt}
\begin{tcolorbox}[
    colback=white,
    colframe=black,
    boxrule=0.5pt,
    arc=0pt,
    left=4pt,
    right=4pt,
    top=3pt,
    bottom=3pt
]
\centering
\small
© 2026 IEEE. Personal use of this material is permitted. Permission from IEEE must be obtained for all other uses, including \\
reprinting/republishing this material for advertising or promotional purposes, collecting new collective works for resale or \\
redistribution to servers or lists, or reuse of any copyrighted component of this work in other works.
\end{tcolorbox}
\end{minipage}
}
}
}

\title{Vision-Language Model Reasoning for Contextual Semantic Mapping in Intralogistics \\
% {\footnotesize \textsuperscript{*}Note: Sub-titles are not captured for https://ieeexplore.ieee.org  and
% should not be used}
\thanks{This research was funded by the InvestBW Innovation Program of the Ministry of Economic Affairs, Labor and Tourism of Baden-Württemberg, Germany.}
}

\author{
\IEEEauthorblockN{Marvin Rüdt$^1$$^,$$^2$}
\IEEEauthorblockA{
marvin.ruedt@kit.edu}
\and
\IEEEauthorblockN{Hao Pang$^1$$^,$$^2$}
\IEEEauthorblockA{
hao.pang@kit.edu}
\and
\IEEEauthorblockN{Constantin Enke$^1$}
\IEEEauthorblockA{
constantin.enke@kit.edu}
\and
\IEEEauthorblockN{Zäzilia Seibold$^1$}
\IEEEauthorblockA{
zaezilia.seibold@kit.edu}
\and
\IEEEauthorblockN{Kai Furmans$^1$}
\IEEEauthorblockA{
kai.furmans@kit.edu}
\thanks{$^1$Institute for Material Handling and
Logistics, Karlsruhe Institute of Technology,
Karlsruhe, Germany.}
\thanks{$^{2}$These authors contributed equally.}
}

%\author{Marvin Rüdt$^1$$^2$, Hao Pang$^1$$^2$,  Constantin Enke$^1$, Zäzilia Seibold$^1$, and Kai Furmans$^1$
%\thanks{$^1$ Authors are with the Institute for Material Handling and Logistics,
%Karlsruhe Institute of Technology, Karlsruhe, Germany.%
%\thanks{$^2$Authors contributed equally}%
%}

%\author{
%\IEEEauthorblockN{Marvin Rüdt\IEEEauthorrefmark{1}, Hao Pang\IEEEauthorrefmark{1}, Constantin Enke, Zäzilia Seibold, and Kai Furmans}
%\IEEEauthorblockA{Institute for Material Handling and Logistics,
%Karlsruhe Institute for Technology, Karlsruhe, Germany\\
%Email: \{marvin.ruedt, hao.pang, constantin.enke, zaezilia.seibold, kai.furmans\}@kit.edu\\
%\IEEEauthorrefmark{1}Authors contributed equally}
%}

\maketitle

\IEEEcopyrightbox

\begin{abstract}
Autonomous mobile robots operating in intralogistics environments rely
on geometric maps for localization and navigation, but lack semantic
understanding of objects and their contextual properties. We present a contextual semantic
mapping pipeline that combines SLAM-based geometric mapping, SAM-based instance
segmentation, instance clustering, and VLM multi-view
reasoning to produce a contextual semantic map representation encoding geometric structure, object class, and
object movability. By aggregating observations across multiple viewpoints and querying
a VLM in a zero-shot, open-vocabulary setting, the pipeline infers contextual
object properties--here demonstrated through movability--without requiring
task-specific training or predefined object categories. We evaluate three VLMs
under two prompting strategies and conduct a component-wise analysis of the pipeline. The proposed pipeline
achieves 98.93\% mIoU for semantic classification and 89.17\% mAcc for object 
movability estimation. Component analysis identifies VLM reasoning as the primary
bottleneck for contextual understanding and instance clustering as the main
limitation for panoptic performance. The resulting semantic map supports
context-aware filtering and robust navigation in dynamic intralogistics
environments.
\end{abstract}


\section{Introduction}
Autonomous mobile robots (AMRs) are increasingly deployed in indoor environments such as intralogistics systems, where they navigate freely using geometric maps generated by simultaneous localization and mapping (SLAM) \cite{cadenaPresentFutureSimultaneous2016}. These maps encode the spatial structure of the environment and support localization and path planning, but lack semantic information about objects and infrastructure, limiting higher-level task execution and reasoning \cite{armeni3DSceneGraph2019, zhouMultiMap3DMultilevelSemantic2023, alqobaliRealtimeSemanticMap2024}.

This limitation is particularly critical in intralogistics systems, which are dynamic and heterogeneous, containing objects such as shelves, pallets, and vehicles whose states change over time. Current systems often struggle to adapt to such changes and may require frequent remapping \cite{lacknerReviewAutonomousMobile2024}. Semantic maps must therefore go beyond object categories and capture contextual properties relevant to robot operation, such as movability, functional state, or spatial relationships.

Classical semantic mapping approaches integrate semantic perception into SLAM pipelines \cite{martinsExtendingMapsSemantic2020, songMonocularCameraLaser2023, alqobaliRealtimeSemanticMap2024}, but rely on closed-vocabulary models limited to predefined categories. Recent work explores open-vocabulary mapping using vision-language models (VLMs) \cite{chenOpenvocabularyQueryableScene2023, huangVisualLanguageMaps2023}, enabling more flexible object recognition. However, these methods primarily focus on object identification and lack contextual reasoning about object properties \cite{pawarSceneUnderstandingSurvey2019, williamsStructuredGenerativeModels2025}. Approaches integrating foundation models such as contrastive language-image pre-training (CLIP) \cite{radfordLearningTransferableVisual2021} and the segment anything model (SAM) \cite{kirillovSegmentAnything2023} further improve open-vocabulary perception and class-agnostic segmentation, yet most operate on single images and do not reason across multiple viewpoints or infer higher-level semantic attributes.

\begin{figure*}[htbp]
    \centering
    \includegraphics[width=\linewidth]{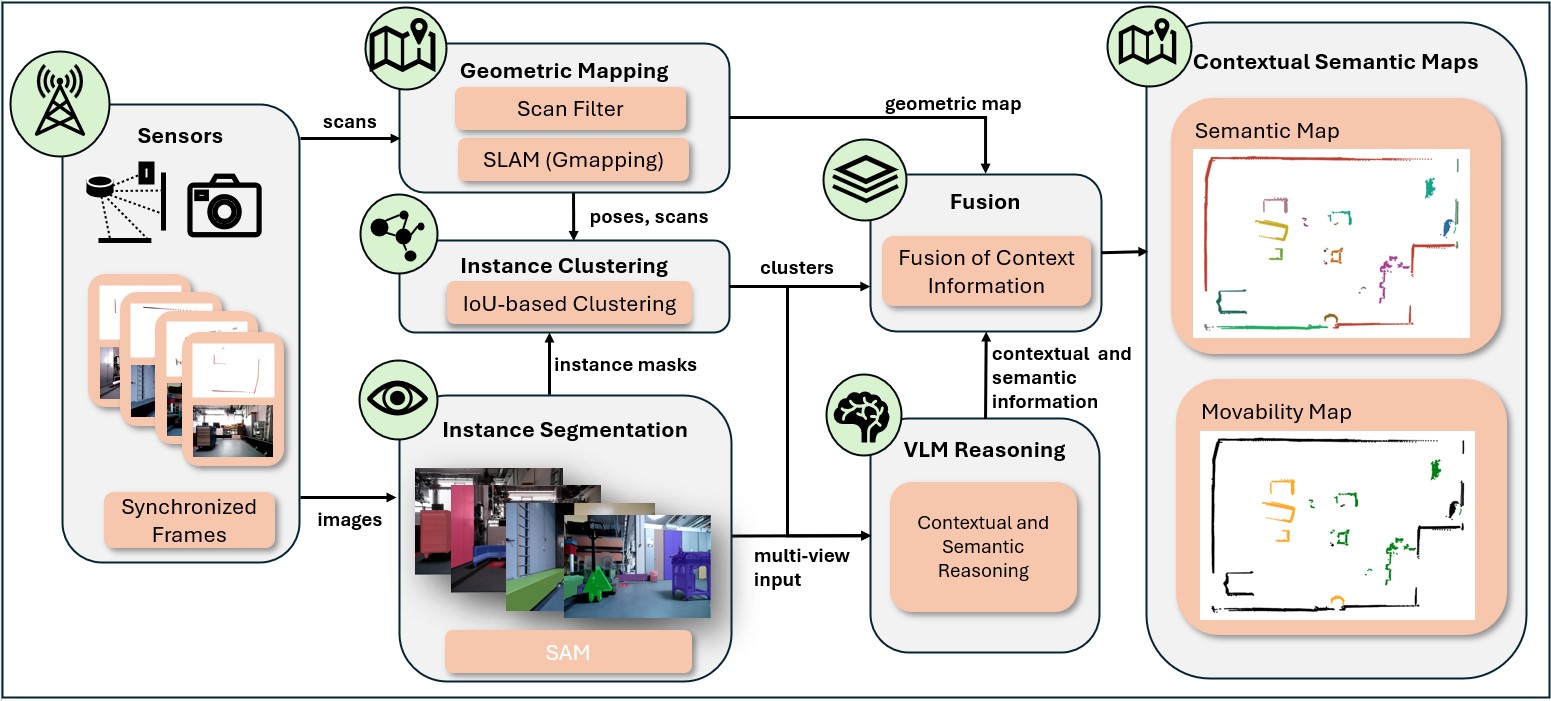}
    \caption{Contextual Semantic Mapping Pipeline}
    \label{fig:mapping_pipeline}
\end{figure*}

To address these gaps, we propose a semantic mapping pipeline that integrates geometric mapping, instance segmentation, instance clustering, and VLM-based contextual reasoning, as illustrated in Fig.~\ref{fig:mapping_pipeline}. RGB images provide visual observations for instance segmentation and serve as input for semantic and contextual inference, while 2D laser scans are used to construct a geometric map via SLAM and to localize and aggregate object instances across multiple frames. A VLM then performs open-vocabulary classification and infers contextual properties--in this work, object movability--enabling the distinction between static infrastructure and potentially dynamic objects. The result is a contextual semantic map representation encoding geometric structure, semantic class, and movability.
The main contributions of this work are:
\begin{itemize}
    \item A modular semantic mapping pipeline integrating SLAM, SAM-based instance segmentation, and VLM-based reasoning for intralogistics environments.
    \item A multi-view reasoning approach enabling robust open-vocabulary object classification and inference of contextual properties such as object movability.
    \item A contextual semantic map representation that augments geometric maps with semantic class and object movability.
    \item A systematic evaluation of individual pipeline components and their contribution to semantic mapping performance.
\end{itemize}

\section{Related Work}
\subsection{Semantic Mapping in Robotics}
Semantic mapping enhances geometric representations by incorporating information at the object and context levels, thereby enabling robots to reason beyond purely spatial structures \cite{raychaudhuriSemanticMappingIndoor2025}. Initial approaches couple SLAM with recognition of defined object classes to assign semantic labels to spatial maps using RGB-D and LiDAR data \cite{alqobaliRealtimeSemanticMap2024, martinsExtendingMapsSemantic2020, songMonocularCameraLaser2023}. For example, \cite{alqobaliRealtimeSemanticMap2024} fuse RGB-D and LiDAR data within a SLAM pipeline and project YOLO-based detections into the map for semantic filtering. Similarly, \cite{martinsExtendingMapsSemantic2020} construct persistent object-level maps, while \cite{songMonocularCameraLaser2023} improve robustness through temporal consistency. While effective, these methods rely on closed-vocabulary models and are therefore limited to predefined object categories.

Several works extend semantic mapping toward multi-level map representations. Hierarchical and object-centric approaches organize environments into multiple abstraction levels \cite{zhouMultiMap3DMultilevelSemantic2023}, while scene graph representations encode relationships between entities \cite{armeni3DSceneGraph2019}. However, these methods primarily focus on structural organization rather than semantic reasoning about object properties and context. In dynamic environments, \cite{zhengSemanticSLAMSystem2025} combine segmentation and motion cues to identify and remove moving objects during SLAM, improving localization accuracy but relying on observed motion.

More recent work explores open-vocabulary semantic mapping. VLMaps \cite{huangVisualLanguageMaps2023} fuse CLIP-based visual-language embeddings with 3D reconstructions, enabling flexible semantic queries at runtime. Related approaches incorporate CLIP features into SLAM pipelines for open-vocabulary object localization and multi-view fusion \cite{chenOpenvocabularyQueryableScene2023, martinsOpenvocabularyOnlineSemantic2025}. While these methods improve flexibility, they primarily focus on object identification and retrieval, and dense feature representations can introduce object ambiguities \cite{huangVisualLanguageMaps2023}.

Overall, existing approaches provide either closed-set semantic maps or open-vocabulary representations without deeper contextual understanding. In contrast, we incorporate vision-language-based reasoning into the mapping process to infer object properties and enable context-aware map filtering.

\subsection{Foundation Models for Visual Perception}
Recent advances in foundation models have significantly expanded visual perception capabilities by providing complementary mechanisms for segmentation, open-vocabulary recognition, and language grounding. SAM \cite{kirillovSegmentAnything2023} enables high-quality, class-agnostic segmentation by generating object masks from image prompts, with extensions supporting concept-level grounding through language \cite{carionSAM3Segment2025}. In contrast, VLMs such as CLIP \cite{radfordLearningTransferableVisual2021} learn a joint embedding space for images and text, enabling open-vocabulary classification. Grounding DINO \cite{liuGroundingDINOMarrying2025} further extends these capabilities by performing language-guided object detection through cross-modal attention.

These models are often combined into integrated pipelines. Grounded-SAM \cite{renGroundedSAMAssembling2024} first generates bounding boxes from text prompts using Grounding DINO and then refines them into pixel-level masks with SAM, enabling automatic object annotation. While effective, such approaches depend on accurate tagging models, which can be unreliable in cluttered environments. In contrast to prompt-based approaches, we adopt a bottom-up paradigm to eliminate dependence on manual prompts and enhance segmentation quality. We first generate fine-grained masks automatically with SAM and then aggregate them into persistent object-level instances through clustering in the geometric map.

Language-guided perception pipelines further incorporate higher-level reasoning into detection. \cite{patzoldLeveragingVisionlanguageModels2025} generate structured scene descriptions using VLMs to guide open-vocabulary detection and segmentation, while \cite{shiVISOGraspVisionlanguageInformed2025} use a VLM in combination with Grounding DINO and SAM for extracting open-vocabulary 3D object representations. However, these approaches rely on single-view reasoning or predefined targets in structured environments, which limits robustness in industrial settings.

Despite these advances, most methods remain image-centric and focus on object identity rather than contextual reasoning and scene understanding. \cite{zemskovaSegmATRonEmbodiedAdaptive2025} incorporate temporal information to improve segmentation quality, but likewise without accounting for context. In contrast, we use segmentation as an intermediate representation, aggregate object observations across multiple views, and enable subsequent semantic reasoning within a spatial map.

\subsection{Language-Based Semantic Reasoning}
Vision-language models enable reasoning beyond object recognition, including the inference of attributes, relationships, and contextual properties \cite{radfordLearningTransferableVisual2021, wangVisionLLMLargeLanguage2023a}. Recent work extends these capabilities to spatial reasoning by incorporating depth and region-level representations \cite{chengSpatialRGPTGroundedSpatial2024}.

In robotics, language models have been applied to perception, mapping, and planning. For example, \cite{jiaoRealtimeIndoorObject2025} incorporate language priors into SLAM, while \cite{shiLLMguidedZeroshotVisual2025} use language-based reasoning for navigation. Structured representations such as scene graphs enable relational reasoning \cite{puigjanerRelationshipawareHierarchical3D2026, xieDSMConstructingDiverse2026}, and systems such as osmAG-LLM \cite{xieOsmAGLLMZeroshotOpenvocabulary2026} performs reasoning over semantic maps to generate navigation goals verified through perception.

Generative approaches further demonstrate the potential of language models for scene understanding. MetaScenes \cite{yuMetaScenesAutomatedReplica2025} combine SAM-based segmentation with VLMs to generate detailed object descriptions, including attributes such as material and physical properties. However, these approaches focus on scene interpretation or reconstruction and do not integrate semantic reasoning into navigation-relevant mapping.

A key challenge in applying language models is robustness. Prior work shows that performance depends on prompting strategies and can suffer from hallucinations \cite{moncada-ramirezAgenticWorkflowsImproving2025, mohantyFutureMLLMPrompting2025}. Moreover, most systems apply language models either at the image level or as post-processing on pre-built representations, often relying on sparse or single-view observations. In contrast, we integrate language-based reasoning directly into the mapping pipeline, using aggregated multi-view observations to infer object properties and enrich the semantic map representation.

\section{Problem Formulation}
\label{sec:problem_formulation}
AMRs operating in indoor environments rely on geometric maps--typically 2D static occupancy grid maps--generated by SLAM algorithms. These maps represent the spatial structure of the environment and can be used for localization and path planning, but do not contain information about the semantic identity or contextual properties of objects present in the scene. The goal of this work is to generate semantic maps of intralogistics environments that augment geometric structure with context-aware object understanding, including properties such as object movability. Such maps enable higher-level operations and natural language queries, for example \textit{"transport the pallet from the transfer station to the shelf"}, which require semantic object understanding and spatial grounding in the environment, while identifying dynamic objects in the environment can improve the robustness of the localization.

\subsection{Sensor Observations}
We employ a standard sensor configuration in industrial mobile robotics: two 2D laser scanners providing 360° range measurements and a forward-facing RGB camera that captures visual information. During exploration of the environment, the robot collects a sequence of synchronized sensor observations

\begin{equation}
O = \{(I_i, S_i)\}_{i=1}^{T}
\end{equation}

where $I_i$ and $S_i$ denote the RGB image and 2D laser scan in the robot coordinate frame at time step $i$, referred to as a \textit{frame}, and $T$ is the total number of frames.

\subsection{Geometric Map Representation}
The geometric map is represented as a 2D point cloud, which serves as a flexible intermediate form convertible to occupancy grid maps or cost maps. The geometric map is a collection of 2-dimensional points:

\begin{equation}
    GM = \{p_i\}_{i=1}^{n}
\end{equation}

where $p_i \in \mathbb{R}^2$ denotes the position $(x, y)$ of point $i$ in the map coordinate frame, and $n$ is the total number of points.

\subsection{Contextual Semantic Map Representation}
The contextual semantic map representation extends $GM$ by attaching per-point semantic and contextual attributes. The resulting representation of the environment is a collection of 6-dimensional points:

\begin{equation}
    SM = \{(p_i, cls_i, ins_i, m_i)\}_{i=1}^{n}
\end{equation}

where $cls_i$ is the semantic object class, $ins_i$ is the instance identifier, and $m_i \in \{immovable, movable, mobile\}$ is the movability property of the corresponding object. We define three movability categories: \textit{immovable} for fixed infrastructure, \textit{movable} for objects that are transportable but lack intrinsic mobility, and \textit{mobile} for objects that are self-moving or mounted on a mobile carrier.

\subsection{Contextual Semantic Mapping Task}
Given the observation sequence $O$, the objective is to estimate $SM$--a geometrically grounded, semantically enriched representation of the environment. For each point $p_i$, the pipeline assigns $cls_i$, $ins_i$, and $m_i$ by combining SAM-based instance segmentation, instance clustering, and VLM-based contextual and semantic reasoning.

\section{Contextual Semantic Mapping Pipeline}
\label{sec:pipeline}
We propose a multimodal mapping framework that integrates geometric mapping, instance-level perception, and VLM-based multi-view reasoning to generate a contextual semantic map representation of intralogistics environments. As illustrated in Fig.~\ref{fig:mapping_pipeline}, the system consists of five main modules: geometric mapping, instance segmentation, instance clustering, VLM-based reasoning, and map fusion.

The system takes as input synchronized measurements from an RGB camera and two 2D laser scanners. Temporal synchronization establishes a point-to-pixel correspondence that associates geometric observations from the laser scanners with visual information extracted from camera images, enabling joint spatial-semantic processing throughout the pipeline.

\subsection{Geometric Mapping}
The laser scans are first processed in a geometric mapping module. We apply a scan filtering step to remove overlapping projected points caused by the spatial offset between the camera and laser scanner. Specifically, when scan points are reprojected into the camera coordinate frame, sensor placement differences can cause multiple points to overlap in the image plane; these ambiguities are filtered prior to downstream processing. After preprocessing, filtered scans are used for 2D geometric mapping using SLAM. In this work, we employ GMapping to estimate robot poses and build the geometric map $GM$ of the environment.

\subsection{Instance Segmentation}
We apply the automatic mask generation pipeline of SAM \cite{kirillovSegmentAnything2023} to each RGB image independently, producing
fine-grained, class-agnostic segmentation masks. The established point-to-pixel correspondence is used to project segmented instances into the geometric map coordinate frame, associating each mask with a spatial footprint in $GM$.

\subsection{Instance Clustering}
To associate observations of the same physical object across frames, we introduce an instance clustering module operating in the geometric map space. Projected instances are clustered based on pairwise intersection-over-union (IoU): instances with $IoU > 0.3$ are grouped as observations of the same object. The resulting clusters serve two purposes--they provide persistent object-level representations for map fusion, and aggregate multi-view observations passed to the VLM reasoning module.

\subsection{VLM-Based Reasoning}

Aggregated multi-view observations of each object cluster are passed to a VLM, which infers object class and movability in a zero-shot, open-vocabulary setting. The reasoning module is described in detail in Section~\ref{sec:vlm_reasoning}.

\subsection{Map Fusion}
The fusion module integrates three information sources: (i) the geometric map $GM$ produced by SLAM, (ii) clustered object instances from segmentation and clustering, and (iii) semantic and contextual attributes inferred by the VLM, to produce the final semantic map $SM$ as defined in Section~\ref{sec:problem_formulation}. Each object instance is represented by its spatial footprint, semantic class, and movability, yielding a contextual semantic map representation that supports both geometric localization and context-aware reasoning.

\section{Vision-Language Model Reasoning}
\label{sec:vlm_reasoning}
To obtain semantic understanding of the environment, we employ a VLM in a zero-shot, open-vocabulary setting to infer object classes and movability--the primary contextual property for distinguishing static infrastructure from potentially dynamic objects in intralogistics environments.

\subsection{Visual Input Representation}

\begin{figure}[htbp]
    \centering
    \includegraphics[width=\linewidth]{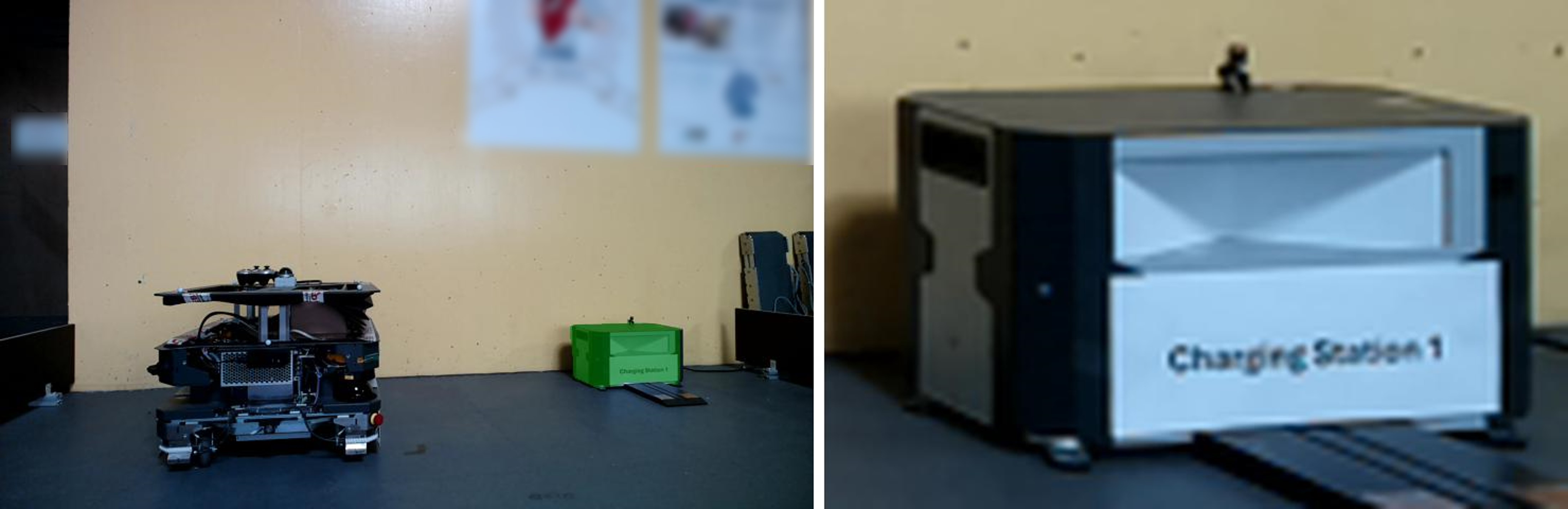}
    \caption{Composite VLM input image showing a charging station with a textual label. Left: full-scene view with green segmentation mask highlight. Right: cropped object view.}
    \label{fig:VLM_Image_Input_CS}
\end{figure}

Each object instance is represented by a composite image comprising (i) a full-scene view in which the target object is highlighted using a green segmentation mask to provide spatial context, and (ii) a cropped close-up of the object that preserves fine-grained visual details (Fig.~\ref{fig:VLM_Image_Input_CS}). The close-up views are generated directly from the segmentation masks with a small amount of padding around the object. To improve robustness to occlusions and partial visibility, multiple views of the same object instance are provided jointly, allowing the VLM to integrate complementary observations across viewpoints and resolve ambiguities that may arise from individual frames.

\subsection{Prompt Design and Reasoning Strategy}

Inference is guided by a structured prompt defining the task, a movability ontology, and a JSON output format. We compare two prompting strategies:
direct JSON prompting, which returns a structured output directly, and CoT-augmented JSON prompting, which precedes the JSON output with an explicit
chain-of-thought (CoT) reasoning process covering visual feature extraction, multi-view aggregation, and reasoning. The effect of CoT augmentation on classification performance is evaluated in Section~\ref{subsec:vlm_studies}. The prompt follows a fixed structure; an abbreviated version is shown below.

\begin{tcolorbox}[colback=gray!5,colframe=gray!50,title=Prompt Template (Abbreviated)]
\small
\textbf{Role:} Expert AI for visual perception in intralogistics\\
\textbf{Task:} Classify the highlighted object and infer its movability\\
\textbf{Input:} Composite images (full-scene + cropped close-up) of the same object instance

\textbf{Process (optional):}
\begin{enumerate}
    \item \textbf{Perception:} Extract visual features (shape, ground contact,
    carriers, context)
    \item \textbf{Multi-view aggregation:} Combine observations and resolve
    ambiguities
    \item \textbf{Reasoning:} Infer object class and movability
\end{enumerate}

\textbf{Movability ontology:}
\begin{enumerate}
    \item \textbf{Immovable:} Fixed infrastructure (e.g., wall, shelf, barrier,
    charging station)
    \item \textbf{Movable:} Transportable, no intrinsic mobility (e.g.,
    pallet, box, container)
    \item \textbf{Mobile:} Self-moving, on wheels, or on mobile carrier (e.g., human,
    AMR, pallet on forklift)
\end{enumerate}

\textbf{Output:} JSON, optionally preceded by CoT reasoning
\end{tcolorbox}

\begin{figure}[htbp]
    \centering
    \includegraphics[width=\linewidth]{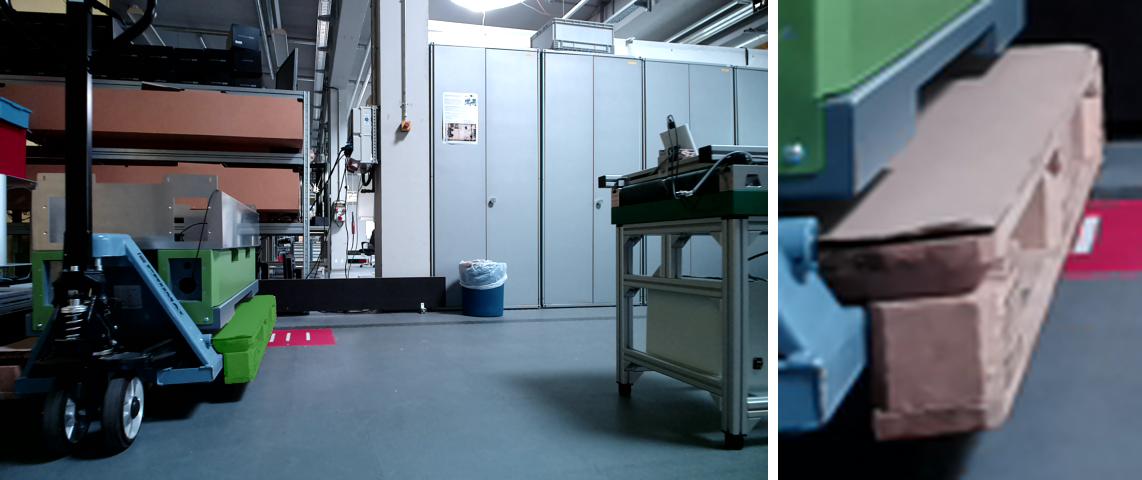}
    \caption{Composite VLM input showing a pallet on a pallet jack. Left: full-scene view with green segmentation mask highlight. Right: cropped object view.}
    \label{fig:VLM_Image_Input_PoPJ}
\end{figure}

Contextual reasoning plays a key role: for example, a pallet is classified as \textit{movable} when standing on the floor, but as \textit{mobile} when placed on a pallet jack or forklift (Fig.~\ref{fig:VLM_Image_Input_PoPJ}). Distinctions like these require reasoning about object relationships, rather than appearance alone. 

\subsection{Structured Output and Map Integration}

The VLM returns a structured JSON containing the inferred object class, object movability, and a short explanation for traceability, ensuring consistent downstream processing:

\begin{tcolorbox}[colback=gray!5,colframe=gray!50,title=JSON Output for
Fig.~\ref{fig:VLM_Image_Input_PoPJ}]
\ttfamily\small
\{\\
\quad "object\_class": "pallet on pallet jack",\\
\quad "movability": mobile,\\
\quad "short\_explanation": "The pallet is currently being transported by a mobile pallet jack, making it mobile in this context."\\
\}
\end{tcolorbox}

If no reliable prediction is possible, a fallback (\texttt{object\_class: unknown}, \texttt{movability: unknown}) is returned.

The predicted attributes $cls_i$ and $m_i$ are assigned to the corresponding object instance and propagated to all map points $p_i$ belonging to that instance, yielding the semantic map $SM$ as defined in Section~\ref{sec:problem_formulation}.

\subsection{Design Choices}
\label{subsec:vlm_design}

The following design choices were evaluated to obtain a robust configuration for VLM-based semantic reasoning.

\textbf{Input representation.}
Explored formats included full-scene images with bounding-box or segmentation mask overlays, cropped-only views, and composite representations. The composite format--combining a full-scene segmentation mask with a cropped object view--yielded the most reliable results. It provides both spatial context and fine-grained object detail, and proved essential for directing model attention toward the highlighted object rather than visually dominant elements.

\textbf{Prompt design.}
Evaluated variants included short and detailed instructions, prompts with and without output examples, and prompts with and without explicit reasoning steps. Best results were obtained with structured prompts containing an explicit movability ontology. Output examples were omitted to avoid anchoring bias toward specific object classes. The effect of CoT-reasoning is analyzed in Section~\ref{subsec:vlm_studies}.

\textbf{Contextual text.}
For process-specific infrastructure such as charging and transfer stations, textual labels were placed on objects in the scene. The VLM reliably leverages such in-scene text even under low resolution and motion blur.

\section{Experimental Results and Evaluation}
\subsection{Experimental Setup}
Experiments are conducted in a controlled intralogistics environment comprising 18 object instances from 13 semantic classes, where instances refer to physical
objects and classes to their semantic categories. Data is recorded using the sensor configuration described in Section~\ref{sec:pipeline} and subsampled
via motion thresholds (30\,cm translation or 15$^\circ$ rotation), yielding 74 frames.\footnote{The dataset, including RGB images,
laser scans, instance annotations, and the VLM prompts, is publicly available: \newline
\url{https://doi.org/10.35097/qx7b62vnbercxzj9}.} Each object is observed from multiple viewpoints; all available views are provided jointly as composite images in a single query.

In Section~\ref{subsec:vlm_studies}, we first conduct a VLM study in which the pipeline is kept fixed and only the semantic inference stage is varied. We analyze the effect of (i) VLM choice and (ii) prompting strategy--direct JSON prompting vs.\ CoT-augmented JSON prompting--on open-vocabulary classification and movability inference accuracy. Three VLMs are evaluated: \textit{Gemini 3.1 Flash Lite} (released March 2026), a lightweight model optimized for efficiency; \textit{GPT-5.1} (released November 2025), a high-capacity general-purpose model; and \textit{GPT-o3} (released April 2025), a reasoning-optimized model designed for structured multi-step inference. This selection covers the trade-off between inference efficiency, model capacity, and reasoning ability. 

To identify system bottlenecks and quantify individual component contributions, we further conduct a component analysis in Section~\ref{subsec:component_analysis} using ground-truth substitution, with the VLM configuration fixed to the best-performing combination from the VLM study. Ground-truth annotations were manually generated for the final map as well as for each intermediate module. For the instance segmentation module, ground-truth instance masks were obtained by manually providing positive and negative point prompts to SAM. For the instance clustering module, objects were manually annotated with bounding boxes, and segmented instances belonging to the same object were explicitly associated. Semantic and movability labels were added to the objects to obtain the groud-truth for the VLM-reasoning module and the contextual semantic maps.

\subsection{Evaluation Metrics}
\textbf{Semantic segmentation} is evaluated using mean intersection over union (mIoU). For each semantic class $c$ present in the ground truth, IoU is computed as 
\begin{equation}
    \text{IoU}_c = \frac{\text{TP}_c}{\text{TP}_c + \text{FP}_c + \text{FN}_c},
\end{equation}
where $\text{TP}_c$, $\text{FP}_c$, and $\text{FN}_c$ denote the number of points correctly classified, falsely assigned to class $c$, and missed for class $c$, respectively. The mIoU is averaged over all classes present in the ground truth.

We evaluate \textbf{movability classification} as balanced per-class accuracy over the three movability classes: immovable, movable, and mobile. For each movability class present in the ground truth, we compute the fraction of valid points of that class whose predicted movability matches the ground-truth. The final score (mAcc) is the average of these per-class accuracies. This gives equal weight to each movability class and avoids domination by classes containing many more points.

Since our mapping pipeline is also able to understand the environment in a panoptic manner (a combination of instance segmentation and semantic segmentation), we further evaluate the 
\textbf{panoptic segmentation} performance using Panoptic Quality (PQ)
\cite{kirillovPanopticSegmentation2019}. Panoptic Quality is defined  as
\begin{equation}
\text{PQ} = \frac{\sum_{(p,g) \in TP} \text{IoU}(p,g)}{|TP| + \frac{1}{2}|FP| + \frac{1}{2}|FN|}.
\end{equation}
 Here, a predicted segment $p$ and a ground-truth segment $g$ are considered a true positive (TP) if their $\text{IoU}(p,g) > 0.5$. PQ is computed per class and then averaged over all classes to obtain the final score.

\subsection{Vision-Language Model Study}
\label{subsec:vlm_studies}

\begin{table}[t]
\caption{Semantic classification (mIoU) and movability accuracy (mAcc) in \%
for different VLM models and prompting strategies. Deviations exceeding 2\%
relative to the direct JSON prompting baseline are highlighted
(red: degradation, green: improvement).}
\centering
\setlength{\tabcolsep}{4pt}
\begin{tabular}{l|cc|cc|cc}
\toprule
\textbf{Model} $\rightarrow$ &
 \multicolumn{2}{c|}{\textbf{Gemini 3.1 FL}}
 & \multicolumn{2}{c|}{\textbf{GPT-5.1}}
 & \multicolumn{2}{c}{\textbf{GPT-o3}} \\
\textbf{Prompt} $\downarrow$
 & mIoU & mAcc & mIoU & mAcc & mIoU & mAcc \\
\midrule
Direct JSON
 & 98.93 & 89.17
 & 74.38 & 96.28
 & 88.67 & 93.38 \\
CoT JSON
 & \textcolor{red}{89.43} & \textcolor{green!70!black}{92.56}
 & \textcolor{red}{68.13} & 96.42
 & \textcolor{green!70!black}{92.89} & \textcolor{red}{89.11} \\
\bottomrule
\end{tabular}
\label{tab:vlm_results}
\end{table}

Table~\ref{tab:vlm_results} reports the semantic classification (mIoU) and
movability accuracy (mAcc) for the three VLMs under two prompting strategies.
All models are evaluated zero-shot, without fine-tuning or in-context examples, via their respective APIs. Typical queries comprise $\sim$700 prompt tokens and $\sim$600 image tokens per view, yielding a $\sim$75-token JSON response; CoT-JSON prompting adds $\sim$300 reasoning tokens. Per-object inference takes approximately 5--10\,s depending on view count and model.

\textbf{VLM comparison.}
Gemini 3.1 Flash Lite achieves the highest semantic classification accuracy
with direct prompting (98.93\% mIoU), while GPT-5.1 lags substantially
(74.38\%), suggesting weaker open-vocabulary visual grounding. GPT-5.1,
however, achieves the highest movability accuracy (96.28\%), indicating
stronger contextual reasoning despite weaker visual classification. GPT-o3
occupies a middle ground on both metrics (88.67\% mIoU, 93.38\% mAcc),
offering a balanced performance across both tasks. Overall, Gemini 3.1 Flash
Lite provides the best trade-off between semantic classification and movability
accuracy.

\textbf{Prompting strategy.}
The effect of CoT reasoning is model-dependent. For
Gemini 3.1 Flash Lite, CoT degrades mIoU by 9.50\% while improving mAcc by
3.39\%, suggesting that explicit reasoning steps introduce classification
uncertainty without proportional benefit for this model. GPT-5.1 shows a
similar pattern, with mIoU decreasing by 6.25\% under CoT. In contrast,
GPT-o3 benefits from CoT reasoning, gaining 4.22\% in mIoU, consistent with
its design as a reasoning-optimized model. Overall, direct JSON prompting
achieves higher semantic classification accuracy across all models, while
movability accuracy remains largely unaffected by the prompting strategy.

\subsection{Component Analysis}
\label{subsec:component_analysis}
Based on the results of Section~\ref{subsec:vlm_studies}, Gemini 3.1 Flash Lite with direct JSON prompting achieves the best overall semantic classification performance and is therefore used for all subsequent experiments. The full pipeline serves as the baseline. To investigate the benefit of multi-view reasoning, we additionally evaluate a single-view variant in which the VLM receives only one composite image per object instance. To identify system bottlenecks, oracle experiments are conducted in which each module is individually replaced with ground-truth inputs. Results are reported in Table~\ref{tab:oracle_experiment}.

\begin{table}[ht]
\caption{Component analysis with oracle substitution. $\checkmark$ and $\times$ indicate enabled and disabled modules, respectively. $\star$ denotes oracle substitution with ground truth. Changes exceeding 2\% relative to the baseline (first row) are highlighted (red: degradation, green: improvement).}
\centering
\begin{tabular}{cccc|cccc}
\toprule
 SAM & multi-view  & VLM & fusion & mIoU & mAcc & PQ \\
\midrule
%\midrule
%\rowcolor{gray!15}
  $\checkmark$ &    $\checkmark$    &  $\checkmark$   &   $\checkmark$    &   98.93   &  89.17  &  89.78  \\
 $\checkmark$  &  $\times$      &  $\checkmark$   &   $\checkmark$    &  \textcolor{red}{65.90}    &  \textcolor{green!70!black}{94.90}  & \textcolor{red}{52.83} \\
%\rowcolor{gray!15}
 $\star$ &    $\checkmark$     &  $\checkmark$   &    $\checkmark$   &   \textcolor{red}{91.74}   &  \textcolor{green!70!black}{96.85}  &   88.59  \\

 $\checkmark$   &  $\star$ &  $\checkmark$   &  $\checkmark$     &   99.81   &  88.15  &  89.08  \\
%\rowcolor{gray!15}   
  $\checkmark$  &   $\checkmark$     &  $\star$  &   $\checkmark$    &    99.48  &   \textcolor{green!70!black}{99.82} & 90.08  \\

 $\checkmark$    &   $\checkmark$      &    $\checkmark$ &  $\star$    &  98.96    &  89.22 &  \textcolor{green!70!black}{98.92}   \\
%\rowcolor{gray!15}  
 $\star$ &  $\star$ &  $\star$ &  $\star$    &   99.91   &  \textcolor{green!70!black}{100}  &  \textcolor{green!70!black}{99.91}   \\
\bottomrule
\end{tabular}
\label{tab:oracle_experiment}
\end{table}
\begin{figure*}[hbtp]
    \centering
    \includegraphics[width=\linewidth]{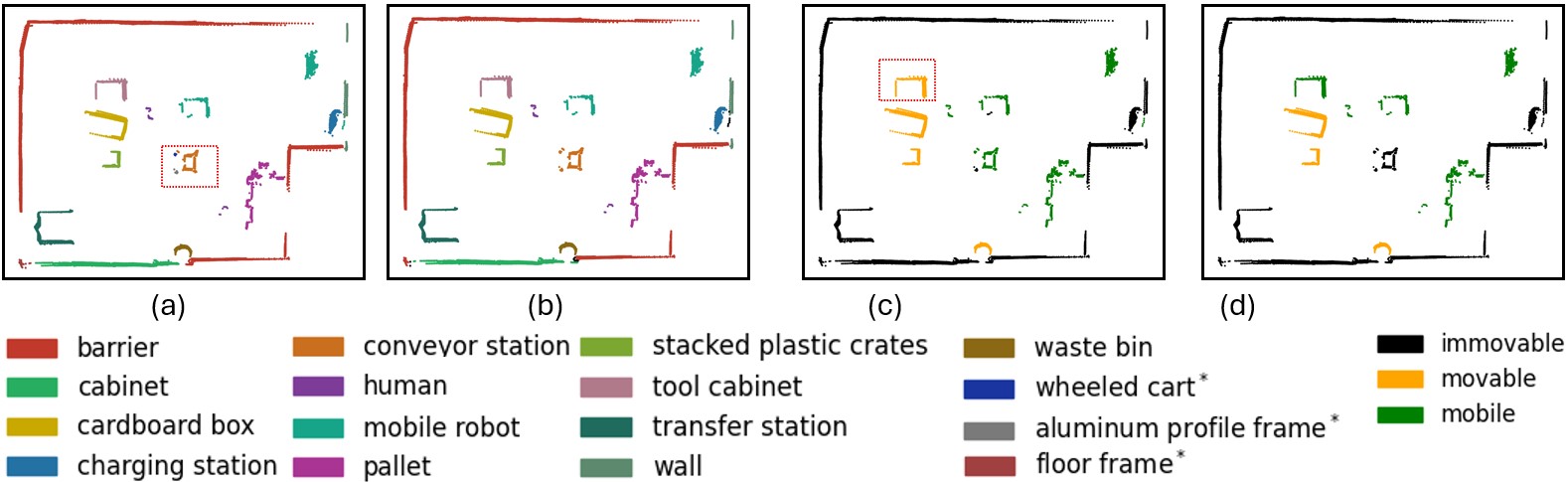}
    \caption{Visualization of the semantic and movability maps from the experiment. (a) Predicted semantic map; (b) ground-truth semantic map; (c) predicted movability map; and  (d) ground-truth movability map. Major incorrect estimations in both predicted maps are highlighted with red bounding boxes. 
Semantic labels marked with * denote categories absent from the ground truth but appearing in the estimated map.}
    \label{fig:map_vis4}
\end{figure*}
The full pipeline achieves 98.93\% mIoU for the semantic segmentation, 89.17\% mAcc for the movability estimation, and 89.78\% PQ for the panoptic segmentation, demonstrating robust scene understanding ability in cluttered intralogistics environments. Qualitative mapping results are shown in Fig.~\ref{fig:map_vis4}, together with the corresponding manually annotated ground-truth for comparison. 
%The major incorrect estimations are highlighted by bounding boxes in the predicted maps. The incorrect semantic prediction for the conveyor station (Fig.\ref{fig:conveyer_station}) is caused by incorrect instance clustering, where two aluminum profiles of this station are not grouped with the main part together, and reasoning on the profiles leads to incorrect semantic prediction: aluminum profile frame and wheeled cart. This indicates the bottleneck of clustering objects in a 2D plane. The tool cabinet (Fig.~\ref{fig:tool_cabinet}) is predicted as movable. According to our definition it should be classified as mobile due to the mounted wheels.

Disabling the multi-view input leads to a slight increase of 5.73\% in the movability mAcc, but causes a substantial drop (over 30\%) in all other metrics, indicating that multi-view input is necessary for maintaining consistent semantic reasoning, while multiple independent reasoning can improve contextual understanding.

Replacing individual components with ground truth provides further insights. Using ground-truth segmentation results in an improvement of 7.68\% in mAcc but a degradation of 7.49\% in semantic mapping performance (mIoU). This observation appears counterintuitive, as one might expect perfect segmentation masks to improve downstream recognition. One possible explanation is that current VLMs may not rely on masks as strict pixel-level grounding signals. Using the ground-truth clustering to aggregate multi-views of one object together does not lead to noticeable improvements, suggesting that clustering is not a bottleneck for the multi-view reasoning. In contrast, substituting the VLM with ground-truth (VLM GT) improves mAcc for over 10\%, indicating that reasoning remains a limiting factor. Improvements at the panoptic level are mainly observed when replacing the fusion stage, suggesting that instance association is the primary bottleneck for panoptic-level mapping.

Finally, when all components are replaced with ground truth, the pipeline approaches optimal performance. The remaining gap from perfect mIoU is mainly attributed to errors introduced during the fusion of point clouds and image observations. This verifies the design of the proposed semantic mapping pipeline.

\section{Conclusion}
We presented a semantic mapping pipeline for intralogistics environments that
integrates SLAM-based geometric mapping, SAM-based instance segmentation, instance clustering, and VLM-based contextual reasoning. By
aggregating object observations across multiple viewpoints and reasoning over
them in a zero-shot, open-vocabulary setting, the pipeline produces a
contextual semantic map representation encoding geometric structure, semantic class, and
object movability--enabling the distinction between static infrastructure and potentially dynamic objects
without requiring task-specific training or predefined object categories.

Experimental results demonstrate that Gemini 3.1 Flash Lite with direct JSON
prompting achieves the highest semantic classification accuracy (98.93\% mIoU),
while component analysis confirms that multi-view reasoning is essential for
consistent spatial and semantic grouping. VLM-based reasoning is identified as
the primary bottleneck for movability estimation, and instance association as
the main limitation for panoptic performance.

The current system operates offline on pre-recorded data and is evaluated in a
single controlled environment. Future work will focus on online semantic map
updating to handle dynamic object states, extension to 3D environments using
3D LiDAR sensors, and evaluation across diverse intralogistics scenarios.
Incorporating additional contextual properties beyond movability--such as
functional state or spatial relationships--represents a further direction
toward richer, task-aware semantic representations.

\bibliographystyle{IEEEtran}
\bibliography{references}

@article{cadenaPresentFutureSimultaneous2016,
	title = {Past, {Present}, and {Future} of {Simultaneous} {Localization} and {Mapping}: {Toward} the {Robust}-{Perception} {Age}},
	volume = {32},
	copyright = {https://ieeexplore.ieee.org/Xplorehelp/downloads/license-information/IEEE.html},
	issn = {1552-3098, 1941-0468},
	shorttitle = {Past, {Present}, and {Future} of {Simultaneous} {Localization} and {Mapping}},
	doi = {10.1109/TRO.2016.2624754},
	language = {en},
	number = {6},
	urldate = {2026-04-25},
	journal = {IEEE Transactions on Robotics},
	author = {Cadena, Cesar and Carlone, Luca and Carrillo, Henry and Latif, Yasir and Scaramuzza, Davide and Neira, Jose and Reid, Ian and Leonard, John J.},
	year = {2016},
	pages = {1309--1332},
}

@article{xieOsmAGLLMZeroshotOpenvocabulary2026,
	title = {{osmAG}-{LLM}: {Zero}-shot open-vocabulary object navigation via semantic maps and large language models reasoning},
	volume = {11},
	copyright = {https://ieeexplore.ieee.org/Xplorehelp/downloads/license-information/IEEE.html},
	issn = {2377-3766, 2377-3774},
	shorttitle = {{osmAG}-{LLM}},
	doi = {10.1109/LRA.2026.3653280},
	language = {en},
	number = {3},
	urldate = {2026-04-24},
	journal = {IEEE Robotics and Automation Letters},
	author = {Xie, Fujing and Schwertfeger, Sören and Blum, Hermann},
	year = {2026},
	pages = {2426--2433},
}

@article{martinsExtendingMapsSemantic2020,
	title = {Extending maps with semantic and contextual object information for robot navigation: {A} learning-based framework using visual and depth cues},
	volume = {99},
	shorttitle = {Extending {Maps} with {Semantic} and {Contextual} {Object} {Information} for {Robot} {Navigation}},
	doi = {10.48550/ARXIV.2003.06336},
	urldate = {2026-02-16},
	journal = {Journal of Intelligent \& Robotic Systems},
	author = {Martins, Renato and Bersan, Dhiego and Campos, Mario F. M. and Nascimento, Erickson R.},
	year = {2020},
	pages = {555--569},
}

@inproceedings{zhouMultiMap3DMultilevelSemantic2023,
	address = {Birmingham, United Kingdom},
	title = {{MultiMap3D}: {A} multi-level semantic perceptual map construction based on {SLAM} and point cloud detection},
	copyright = {https://doi.org/10.15223/policy-029},
	isbn = {979-8-3503-3585-9},
	shorttitle = {{MultiMap3D}},
	urldate = {2026-02-16},
	booktitle = {2023 28th {International} {Conference} on {Automation} and {Computing} ({ICAC})},
	publisher = {IEEE},
	author = {Zhou, Jingyang and Elksnis, Arturs and Fu, Zhiquan and Chen, Bai and Yang, Chenguang},
	year = {2023},
	pages = {1--6},
}

@inproceedings{shiVISOGraspVisionlanguageInformed2025,
	address = {Hangzhou, China},
	title = {{VISO}-{Grasp}: {Vision}-language informed spatial object-centric 6-{DoF} active view planning and grasping in clutter and invisibility},
	copyright = {https://doi.org/10.15223/policy-029},
	isbn = {979-8-3315-4393-8},
	shorttitle = {{VISO}-{Grasp}},
	doi = {10.1109/IROS60139.2025.11246329},
	language = {en},
	urldate = {2026-04-24},
	booktitle = {2025 {IEEE}/{RSJ} {International} {Conference} on {Intelligent} {Robots} and {Systems} ({IROS})},
	publisher = {IEEE},
	author = {Shi, Yitian and Wen, Di and Chen, Guanqi and Welte, Edgar and Liu, Sheng and Peng, Kunyu and Stiefelhagen, Rainer and Rayyes, Rania},
	year = {2025},
	pages = {14931--14938},
}

@inproceedings{huangVisualLanguageMaps2023,
	address = {London, United Kingdom},
	title = {Visual language maps for robot navigation},
	copyright = {https://doi.org/10.15223/policy-029},
	isbn = {979-8-3503-2365-8},
	doi = {10.1109/ICRA48891.2023.10160969},
	language = {en},
	urldate = {2026-04-24},
	booktitle = {2023 {IEEE} {International} {Conference} on {Robotics} and {Automation} ({ICRA})},
	publisher = {IEEE},
	author = {Huang, Chenguang and Mees, Oier and Zeng, Andy and Burgard, Wolfram},
	year = {2023},
	pages = {10608--10615},
}

@inproceedings{wangVisionLLMLargeLanguage2023a,
	title = {{VisionLLM}: {Large} language model is also an open-ended decoder for vision-centric tasks},
	volume = {36},
	booktitle = {Advances in {Neural} {Information} {Processing} {Systems}},
	publisher = {Curran Associates, Inc.},
	author = {Wang, Wenhai and Chen, Zhe and Chen, Xiaokang and Wu, Jiannan and Zhu, Xizhou and Zeng, Gang and Luo, Ping and Lu, Tong and Zhou, Jie and Qiao, Yu and Dai, Jifeng},
	editor = {Oh, A. and Naumann, T. and Globerson, A. and Saenko, K. and Hardt, M. and Levine, S.},
	year = {2023},
	pages = {61501--61513},
}

@article{mohantyFutureMLLMPrompting2025,
	title = {The future of {MLLM} prompting is adaptive: a comprehensive experimental evaluation of prompt engineering methods for robust multimodal performance},
	copyright = {Creative Commons Attribution Non Commercial No Derivatives 4.0 International},
	shorttitle = {The {Future} of {MLLM} {Prompting} is {Adaptive}},
	doi = {10.48550/ARXIV.2504.10179},
	journal = {Transactions on Machine Learning Research},
	author = {Mohanty, Anwesha and Parthasarathy, Venkatesh Balavadhani and Shahid, Arsalan},
	year = {2025},
}

@article{williamsStructuredGenerativeModels2025,
	title = {Structured generative models for scene understanding},
	volume = {133},
	issn = {0920-5691, 1573-1405},
	doi = {10.1007/s11263-024-02316-z},
	language = {en},
	number = {5},
	urldate = {2026-04-08},
	journal = {International Journal of Computer Vision},
	author = {Williams, Christopher K. I.},
	year = {2025},
	pages = {2845--2867},
}

@article{chengSpatialRGPTGroundedSpatial2024,
	title = {{SpatialRGPT}: {Grounded} spatial reasoning in vision-language models},
	language = {en},
	journal = {38th Conference on Neural Information Processing Systems},
	author = {Cheng, An-Chieh and Yin, Hongxu and Fu, Yang and Guo, Qiushan and Yang, Ruihan and Kautz, Jan and Wang, Xiaolong and Liu, Sifei},
	year = {2024},
	pages = {135062 -- 135093},
}

@article{zhengSemanticSLAMSystem2025,
	title = {Semantic {SLAM} system for mobile robots based on large visual model in complex environments},
	volume = {15},
	issn = {2045-2322},
	doi = {10.1038/s41598-025-90340-5},
	language = {en},
	number = {1},
	urldate = {2026-02-25},
	journal = {Scientific Reports},
	author = {Zheng, Chao and Zhang, Peng and Li, Yanan},
	year = {2025},
	pages = {8450},
}

@misc{raychaudhuriSemanticMappingIndoor2025,
	title = {Semantic mapping in indoor embodied {AI}: {A} survey on advances, challenges, and future directions},
	doi = {10.48550/arXiv.2501.05750},
	language = {en},
	urldate = {2026-02-26},
	publisher = {arXiv},
	author = {Raychaudhuri, Sonia and Chang, Angel X.},
	year = {2025},
	note = {arXiv preprint arXiv:2501.05750},
}

@misc{kirillovSegmentAnything2023,
	title = {Segment anything},
	doi = {10.48550/arXiv.2304.02643},
	language = {en},
	urldate = {2026-03-17},
	publisher = {arXiv},
	author = {Kirillov, Alexander and Mintun, Eric and Ravi, Nikhila and Mao, Hanzi and Rolland, Chloe and Gustafson, Laura and Xiao, Tete and Whitehead, Spencer and Berg, Alexander C. and Lo, Wan-Yen and Dollár, Piotr and Girshick, Ross},
	year = {2023},
	note = {arXiv preprint arXiv:2304.02643},
}

@misc{carionSAM3Segment2025,
	title = {{SAM} 3: {Segment} anything with concepts},
	language = {en},
	publisher = {arXiv},
	author = {Carion, Nicolas and Gustafson, Laura and Hu, Yuan-Ting and Debnath, Shoubhik and Hu, Ronghang and Suris, Didac and Ryali, Chaitanya and Alwala, Kalyan Vasudev and Khedr, Haitham and Huang, Andrew and Lei, Jie and Ma, Tengyu and Kalla, Arpit and Marks, Markus and Greer, Joseph and Wang, Meng and Sun, Peize and Rädle, Roman and Mavroudi, Effrosyni and Xu, Katherine and Wu, Tsung-Han and Zhou, Yu and Momeni, Liliane and Hazra, Rishi and Ding, Shuangrui and Vaze, Sagar and Porcher, Francois and Li, Feng and Li, Siyuan and Kamath, Aishwarya and Kei, Ho and Dollár, Piotr and Ravi, Nikhila and Saenko, Kate and Zhang, Pengchuan and Feichtenhofer, Christoph},
	year = {2025},
	note = {arXiv preprint 	arXiv:2511.16719},
}

@article{zemskovaSegmATRonEmbodiedAdaptive2025,
	title = {{SegmATRon}: {Embodied} adaptive semantic segmentation for indoor environment},
	volume = {638},
	issn = {09252312},
	shorttitle = {{SegmATRon}},
	doi = {10.1016/j.neucom.2025.130169},
	language = {en},
	urldate = {2025-05-09},
	journal = {Neurocomputing},
	author = {Zemskova, Tatiana and Kichik, Margarita and Yudin, Dmitry and Staroverov, Aleksei and Panov, Aleksandr},
	year = {2025},
	pages = {130169},
}

@inproceedings{pawarSceneUnderstandingSurvey2019,
	address = {Jaipur, India},
	title = {Scene understanding: {A} survey to see the world at a single glance},
	copyright = {https://doi.org/10.15223/policy-029},
	isbn = {978-1-7281-1711-9},
	shorttitle = {Scene {Understanding}},
	doi = {10.1109/ICCT46177.2019.8969051},
	urldate = {2026-04-08},
	booktitle = {2019 2nd {International} {Conference} on {Intelligent} {Communication} and {Computational} {Techniques} ({ICCT})},
	publisher = {IEEE},
	author = {Pawar, Prajakta Ganesh and Devendran, V},
	year = {2019},
	pages = {182--186},
}

@misc{puigjanerRelationshipawareHierarchical3D2026,
	title = {Relationship-aware hierarchical {3D} scene graph for task reasoning},
	doi = {10.48550/arXiv.2602.02456},
	language = {en},
	urldate = {2026-03-17},
	publisher = {arXiv},
	author = {Puigjaner, Albert Gassol and Zacharia, Angelos and Alexis, Kostas},
	year = {2026},
	note = {arXiv preprint arXiv:2602.02456},
}

@article{lacknerReviewAutonomousMobile2024,
	title = {Review of autonomous mobile robots in intralogistics: state-of-the-art, limitations and research gaps},
	volume = {130},
	issn = {22128271},
	shorttitle = {Review of autonomous mobile robots in intralogistics},
	doi = {10.1016/j.procir.2024.10.187},
	language = {en},
	urldate = {2026-02-16},
	journal = {Procedia CIRP},
	author = {Lackner, Thorge and Hermann, Julian and Kuhn, Christian and Palm, Daniel},
	year = {2024},
	pages = {930--935},
}

@misc{jiaoRealtimeIndoorObject2025,
	title = {Real-time indoor object {SLAM} with {LLM}-enhanced priors},
	copyright = {arXiv.org perpetual, non-exclusive license},
	doi = {10.48550/ARXIV.2509.21602},
	language = {en},
	urldate = {2026-03-17},
	publisher = {arXiv},
	author = {Jiao, Yang and Qiu, Yiding and Christensen, Henrik I.},
	year = {2025},
	note = {arXiv preprint arXiv:2509.21602},
}

@inproceedings{kirillovPanopticSegmentation2019,
	address = {Long Beach, CA, USA},
	title = {Panoptic segmentation},
	copyright = {https://doi.org/10.15223/policy-029},
	isbn = {978-1-7281-3293-8},
	doi = {10.1109/CVPR.2019.00963},
	urldate = {2026-04-10},
	booktitle = {2019 {IEEE}/{CVF} {Conference} on {Computer} {Vision} and {Pattern} {Recognition} ({CVPR})},
	publisher = {IEEE},
	author = {Kirillov, Alexander and He, Kaiming and Girshick, Ross and Rother, Carsten and Dollár, Piotr},
	year = {2019},
	pages = {9396--9405},
}

@inproceedings{chenOpenvocabularyQueryableScene2023,
	address = {London, United Kingdom},
	title = {Open-vocabulary queryable scene representations for real world planning},
	copyright = {https://doi.org/10.15223/policy-029},
	isbn = {979-8-3503-2365-8},
	doi = {10.1109/ICRA48891.2023.10161534},
	language = {en},
	urldate = {2026-04-24},
	booktitle = {2023 {IEEE} {International} {Conference} on {Robotics} and {Automation} ({ICRA})},
	author = {Chen, Boyuan and Xia, Fei and Ichter, Brian and Rao, Kanishka and Gopalakrishnan, Keerthana and Ryoo, Michael S. and Stone, Austin and Kappler, Daniel},
	year = {2023},
	pages = {11509--11522},
}

@article{martinsOpenvocabularyOnlineSemantic2025,
	title = {Open-vocabulary online semantic mapping for {SLAM}},
	volume = {10},
	copyright = {https://creativecommons.org/licenses/by/4.0/legalcode},
	issn = {2377-3766, 2377-3774},
	doi = {10.1109/LRA.2025.3617736},
	language = {en},
	number = {11},
	urldate = {2026-04-24},
	journal = {IEEE Robotics and Automation Letters},
	author = {Martins, Tomas Berriel and Oswald, Martin R. and Civera, Javier},
	year = {2025},
	pages = {11745--11752},
}

@inproceedings{shiLLMguidedZeroshotVisual2025,
	address = {Munich, Germany},
	title = {{LLM}-guided zero-shot visual object navigation with building semantic map},
	copyright = {https://doi.org/10.15223/policy-029},
	isbn = {979-8-3315-3161-4},
	doi = {10.1109/SII59315.2025.10870978},
	language = {en},
	urldate = {2026-03-17},
	booktitle = {2025 {IEEE}/{SICE} {International} {Symposium} on {System} {Integration} ({SII})},
	author = {Shi, Jin and Yagi, Satoshi and Yamamori, Satoshi and Morimoto, Jun},
	year = {2025},
	pages = {1274--1279},
}

@article{songMonocularCameraLaser2023,
	title = {Monocular camera and laser based semantic mapping system with temporal-spatial data association for indoor mobile robots},
	volume = {82},
	issn = {1380-7501, 1573-7721},
	doi = {10.1007/s11042-023-14796-1},
	language = {en},
	number = {22},
	urldate = {2026-02-25},
	journal = {Multimedia Tools and Applications},
	author = {Song, Xu and Zhijiang, Zuo and Liang, Xuan and Huaidong, Zhou},
	year = {2023},
	pages = {34459--34484},
}

@misc{yuMetaScenesAutomatedReplica2025,
	title = {{MetaScenes}: {Towards} automated replica creation for real-world {3D} scans},
	shorttitle = {{MetaScenes}},
	doi = {10.48550/arXiv.2505.02388},
	language = {en},
	urldate = {2025-05-09},
	publisher = {arXiv},
	author = {Yu, Huangyue and Jia, Baoxiong and Chen, Yixin and Yang, Yandan and Li, Puhao and Su, Rongpeng and Li, Jiaxin and Li, Qing and Liang, Wei and Zhu, Song-Chun and Liu, Tengyu and Huang, Siyuan},
	year = {2025},
	note = {arXiv preprint
arXiv:2505.02388},
}

@article{patzoldLeveragingVisionlanguageModels2025,
	title = {Leveraging vision-language models for open-vocabulary instance segmentation and tracking},
	volume = {10},
	language = {en},
	number = {11},
	journal = {IEEE Robotics and Automation Letters},
	author = {Pätzold, Bastian and Nogga, Jan and Behnke, Sven},
	year = {2025},
	pages = {11578 -- 11585},
}

@article{radfordLearningTransferableVisual2021,
	title = {Learning transferable visual models from natural language supervision},
	volume = {139},
	language = {en},
	journal = {Proceedings of the 38th International Conference on Machine Learning},
	publisher = {Proceedings of Machine Learning Research},
	author = {Radford, Alec and Kim, Jong Wook and Hallacy, Chris and Ramesh, Aditya and Goh, Gabriel and Agarwal, Sandhini and Sastry, Girish and Askell, Amanda and Mishkin, Pamela and Clark, Jack and Krueger, Gretchen and Sutskever, Ilya},
	year = {2021},
	pages = {8748--8763},
}

@article{xieDSMConstructingDiverse2026,
	title = {{DSM}: {Constructing} a diverse semantic map for {3D} visual grounding},
	volume = {11},
	copyright = {https://ieeexplore.ieee.org/Xplorehelp/downloads/license-information/IEEE.html},
	issn = {2377-3766, 2377-3774},
	shorttitle = {{DSM}},
	doi = {10.1109/LRA.2026.3678129},
	language = {en},
	number = {5},
	urldate = {2026-04-24},
	journal = {IEEE Robotics and Automation Letters},
	author = {Xie, Qinghongbing and Liang, Zijian and Li, Fuhao and Zeng, Long},
	year = {2026},
	pages = {6344--6351},
}

@incollection{liuGroundingDINOMarrying2025,
	address = {Cham},
	title = {Grounding {DINO}: {Marrying} {DINO} with grounded pre-training for open-set object detection},
	volume = {15105},
	isbn = {978-3-031-72969-0 978-3-031-72970-6},
	shorttitle = {Grounding {DINO}},
	doi = {10.1007/978-3-031-72970-6_3},
	language = {en},
	urldate = {2026-04-24},
	booktitle = {Computer {Vision} – {ECCV} 2024},
	publisher = {Springer Nature Switzerland},
	author = {Liu, Shilong and Zeng, Zhaoyang and Ren, Tianhe and Li, Feng and Zhang, Hao and Yang, Jie and Jiang, Qing and Li, Chunyuan and Yang, Jianwei and Su, Hang and Zhu, Jun and Zhang, Lei},
	editor = {Leonardis, Aleš and Ricci, Elisa and Roth, Stefan and Russakovsky, Olga and Sattler, Torsten and Varol, Gül},
	year = {2025},
	note = {Series Title: Lecture Notes in Computer Science},
	pages = {38--55},
}

@misc{renGroundedSAMAssembling2024,
	type = {{arXiv} preprint},
	title = {Grounded {SAM}: assembling open-world models for diverse visual tasks},
	shorttitle = {Grounded {SAM}},
	doi = {10.48550/arXiv.2401.14159},
	language = {en},
	urldate = {2025-05-14},
	publisher = {arXiv},
	author = {Ren, Tianhe and Liu, Shilong and Zeng, Ailing and Lin, Jing and Li, Kunchang and Cao, He and Chen, Jiayu and Huang, Xinyu and Chen, Yukang and Yan, Feng and Zeng, Zhaoyang and Zhang, Hao and Li, Feng and Yang, Jie and Li, Hongyang and Jiang, Qing and Zhang, Lei},
	year = {2024},
	note = {arXiv preprint arXiv.2401.14159},
}

@article{moncada-ramirezAgenticWorkflowsImproving2025,
	title = {Agentic workflows for improving large language model reasoning in robotic object-centered planning},
	volume = {14},
	issn = {2218-6581},
	doi = {10.3390/robotics14030024},
	language = {en},
	number = {3},
	urldate = {2026-02-25},
	journal = {Robotics},
	author = {Moncada-Ramirez, Jesus and Matez-Bandera, Jose-Luis and Gonzalez-Jimenez, Javier and Ruiz-Sarmiento, Jose-Raul},
	year = {2025},
	pages = {24},
}

@article{alqobaliRealtimeSemanticMap2024,
	title = {A real-time semantic map production system for indoor robot navigation},
	volume = {24},
	issn = {1424-8220},
	doi = {10.3390/s24206691},
	language = {en},
	number = {20},
	urldate = {2026-02-26},
	journal = {Sensors},
	author = {Alqobali, Raghad and Alnasser, Reem and Rashidi, Asrar and Alshmrani, Maha and Alhmiedat, Tareq},
	year = {2024},
	pages = {6691},
}

@inproceedings{armeni3DSceneGraph2019,
	address = {Seoul, South Korea},
	title = {{3D} scene graph: {A} structure for unified semantics, {3D} space, and camera},
	copyright = {https://ieeexplore.ieee.org/Xplorehelp/downloads/license-information/IEEE.html},
	isbn = {978-1-7281-4803-8},
	shorttitle = {{3D} {Scene} {Graph}},
	doi = {10.1109/ICCV.2019.00576},
	language = {en},
	urldate = {2026-02-16},
	booktitle = {2019 {IEEE}/{CVF} {International} {Conference} on {Computer} {Vision} ({ICCV})},
	author = {Armeni, Iro and He, Zhi-Yang and Zamir, Amir and Gwak, Junyoung and Malik, Jitendra and Fischer, Martin and Savarese, Silvio},
	year = {2019},
	pages = {5663--5672},
}

\end{document}